\documentclass{article}

\usepackage{PRIMEarxiv}

\usepackage[utf8]{inputenc} 
\usepackage[T1]{fontenc}    
\usepackage{hyperref}       
\usepackage{url}            
\usepackage{booktabs}       
\usepackage{amsfonts}       
\usepackage{nicefrac}       
\usepackage{microtype}      
\usepackage{lipsum}
\usepackage{fancyhdr}       
\usepackage{graphicx}       
\graphicspath{{media/}}     

\usepackage{capt-of}
\usepackage[skip=0.5ex]{subcaption}
\usepackage{amsmath}
\usepackage{comment}
\usepackage{floatpag}

\usepackage{xcolor}
\definecolor{RedGraph}{RGB}{254,194,194}
\definecolor{GreenGraph}{RGB}{223,236,210}
\definecolor{PurpleGraph}{RGB}{216,204,226}
\definecolor{BlueGraph}{RGB}{194,216,232}
\definecolor{YellowGraph}{RGB}{245,232,194}
\definecolor{PinkGraph}{RGB}{235,194,214}
\newcommand*{\rom}[1]{\expandafter\@slowromancap\romannumeral #1@}

\usepackage[ruled,vlined]{algorithm2e}

\SetCommentSty{mycommfont}

\newcommand*{\img}[1]{%
    \raisebox{-.2\baselineskip}{%
        \includegraphics[
        height=.8\baselineskip,
        width=.8\baselineskip,
        ]{#1}%
    }%
}

\fancypagestyle{FirstPage}{
\lfoot{\fontsize{8}{12} \selectfont This is the version of the article before peer review or editing, as submitted by an author to Journal of Physics: Conference Series. IOP Publishing Ltd is not responsible for any errors or omissions in this version of the manuscript or any version derived from it. To cite this article, the Version of Record is available online at \href{https://doi.org/10.1088/1742-6596/2265/2/022035}{https://doi.org/10.1088/1742-6596/2265
/2/022035}.}}

\title{Wind Park Power Prediction: Attention-Based Graph Networks and Deep Learning to Capture Wake Losses}

\author{
  Lars Ødegaard Bentsen \\
  Department of Technology Systems (ITS) \\
  University of Oslo \\
  Kjeller, Norway \\
  \texttt{lars.bentsen@its.uio.no} \\
   \And
  Narada Dilp Warakagoda \\
  Department of Technology Systems (ITS) \\
  University of Oslo \\
  Kjeller, Norway \\
     \And
  Roy Stenbro \\
  Institute for Energy Systems (IFE) \\
  Kjeller, Norway \\
     \And
  Paal Engelstad \\
  Department of Technology Systems (ITS) \\
  University of Oslo \\
  Kjeller, Norway \\
}

\begin{document}


\maketitle
\pagenumbering{gobble}
\thispagestyle{FirstPage}
\begin{abstract}
With the increased penetration of wind energy into the power grid, it has become increasingly important to be able to predict the expected power production for larger wind farms. 
Deep learning (DL) models can learn complex patterns in the data and have found wide success in predicting wake losses and expected power production.
This paper proposes a modular framework for attention-based graph neural networks (GNN), where attention can be applied to any desired component of a graph block.
The results show that the model significantly outperforms a multilayer perceptron (MLP) and a bidirectional LSTM (BLSTM) model, while delivering performance on-par with a vanilla GNN model.
Moreover, we argue that the proposed graph attention architecture can easily adapt to different applications by offering flexibility into the desired attention operations to be used, which might depend on the specific application.
Through analysis of the attention weights, it was showed that employing attention-based GNNs can provide insights into what the models learn. 
In particular, the attention networks seemed to realise turbine dependencies that aligned with some physical intuition about wake losses. 
\end{abstract}


\section{Introduction}

Wind energy has been subject to rapid expansion in recent years, with a record 82 GW of new wind capacity installed globally, and a year-on-year growth of 53\% in 2020 \cite{gwec2021}.
Due to the intermittent nature of wind energy, research into wind power modelling and forecasting will remain important to establish wind as a more reliable energy resource.

Turbines have traditionally been characterised by their power curves, a simple relationship that indicates the expected power production at different wind speeds.
However, with the construction of large wind farms, interaction losses from turbines in near vicinity to each other become prominent, resulting in actual production values deviating from power curve predictions.
The wind velocity deficit and increased turbulence generated downstream of a turbine characterise the turbine wake, which result in turbines situated downstream seeing a reduction in the power generated.   
Significant efforts have been spent trying to account for these wake losses, modelling the complicated flow patterns that arise downstream within a farm.
A simple, but widely used method, has been the Jensen wake model, which calculates the velocity deficit downstream of a single wind turbine \cite{jensen1983}.
High resolution computational fluid dynamics can yield high fidelity results \cite{sanderse2011}, but a problem with such numerical methods have been their high computational cost. 

DL-based models have the potential to yield accurate results, while also significantly reduce run times for trained models.
Various studies have reported success in training artificial neural networks (ANN) to predict wind farm power from the free stream wind velocity and direction \cite{manobel2018, yan2019}.
A two-dimensional power curve was introduced by Yan et al. \cite{yan2019}, where the expected power could be determined from the wind speed and direction, thereby taking wake losses into account. 
Ti et al. \cite{ti2020} trained an MLP to predict spatial velocity deficit and added turbulence downstream, aiming to output high resolution flow fields similar to those of numerical methods. 

\pagenumbering{arabic}
\setcounter{page}{2}

Tastu et al. \cite{tastu2013} considered spatial dynamics by using measurements from a number of farms scattered across Denmark, to make their predictions for a target wind farm.
Similarly, both \cite{grover2015deep} and \cite{ghaderi2017deep} proposed spatio-temporal models to make predictions at a number of wind sites jointly. 
Ghaderi et al. \cite{ghaderi2017deep} modelled the spatial characteristics by utilising graphical representations, where the weather stations were represented by nodes and the physical distance between them as edges in a graph, using a Long Short-Term Memory (LSTM) network to make the forecasts. 


Although Recurrent (RNN) and Convolutional Neural Networks (CNN) have predominantly been used for time-series analysis \cite{hanifi2020critical, hong2019hybrid, memarzadeh2020new}, research has also emerged in applying such techniques to capture spatial features.
CNNs have shown superior performance in extracting features from image data.
In \cite{kou2020deep} and \cite{zhu2019learning}, the authors aimed to leverage some of these attributes, by organising the input data in a grid structure, where each pixel contained information for a particular physical location.
Despite yielding good results, a drawback of such an approach is that the strict ordering of the input data might pose some challenges for complex farm layouts.

Graphs have been used to represent data characterised by complex relations, where the data is not naturally ordered in a strict grid-like structure.
Similar to \cite{ghaderi2017deep}, the authors of \cite{chen2020graph}, \cite{khodayar2018spatio} and \cite{stanczyk2021deep} also aimed to improve predictions by considering measurements from different wind farms, but instead employed graph convolutional models.
Bleeg \cite{bleeg2020graph} used a GNN as a surrogate model for a steady-state Reynolds-averaged Navier–Stokes (RANS) model to learn turbine interaction losses within a single wind farm.
Yu et al. \cite{yu2020superposition} constructed a spatio-temporal model to predict wind power for individual turbines using a combination of GNNs and RNNs, trained on data for four offshore wind farms. 
To improve on the vanilla GNN model, Park et al. \cite{park2019physics} proposed the physics-induced GNN (PGNN), where the intensity of different turbine interactions was regulated by computing weights using a physics-induced wake function.
The authors showed the potential benefits of introducing relative weighting of upstream turbines to learn interaction losses, using data simulated in FLORIS \cite{FLORIS_2021}. 
Albeit not fully learned from the data, the weighting introduced in the PGNN draws some resemblance to an attention mechanism, as it regulates the relative impact of neighbouring turbines when predicting the power for a target turbine. 

Attention was first introduced by Bahdanau et al. \cite{bahdanau2014neural}, and has become an integral part of sequence modelling using DL.  
With regards to wind, attention mechanisms have mainly been used to extract temporal tendencies in time-series data, and thereby improve RNN-based forecasting models \cite{fu2019spatiotemporal, li2020short, niu2020wind}. 
Due to the success of attention-based sequence models, there has been a surge in the application of such techniques to other DL domains, such as the development of Graph Attention Networks \cite{chen2021edge, georgousis2021graph, lee2019attention, velivckovic2017graph}.
Nevertheless, these networks have yet to find their way into the wind modelling community, as the authors of this paper cannot report on finding any research discussing the use of attention-based GNNs applied to wind energy. 

The main contributions of this paper can be summarised as follows: 
\begin{itemize}
  \item A BLSTM- and an MLP-model, where inputs are organised around individual turbines, are used to predict the power for a target turbine given the location of upstream neighbours.
  \item We propose a modular graph network architecture with attention mechanisms applied to any desired components of a GNN model. 
  To the knowledge of the authors, this is the first paper that proposes a framework where attention can be applied to any major operation in a GNN. 
  \item Attention weights are investigated to assess the potency of the proposed GNN framework. 
\end{itemize}
This study shows the advantages of a modular GNN-attention framework. 
By analysing various DL-based models, the study also aims to compare the performance of different techniques and show new ways in which DL can be applied to wind power and wake modelling. 

\section{Theory}\label{sec_theory}

\subsection{Bidirectional LSTM}\label{subsec_blstm}
\begin{figure}
\begin{minipage}[t]{0.56\linewidth}
    \centering
    \includegraphics[width=7.8cm]{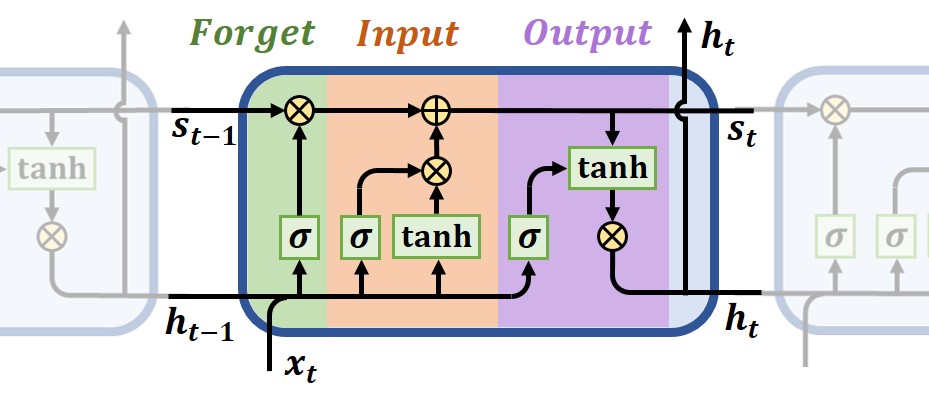}
    \caption[width=15cm]{General structure of an LSTM network. 
    The \textit{Forget, Input} and \textit{Output} notation aims to illustrate how the different gating mechanisms operate.}
    \label{fig:lstm_block}
\end{minipage}%
    \hfill%
\begin{minipage}[t]{0.4\linewidth}
    \centering
    \includegraphics[width=5cm]{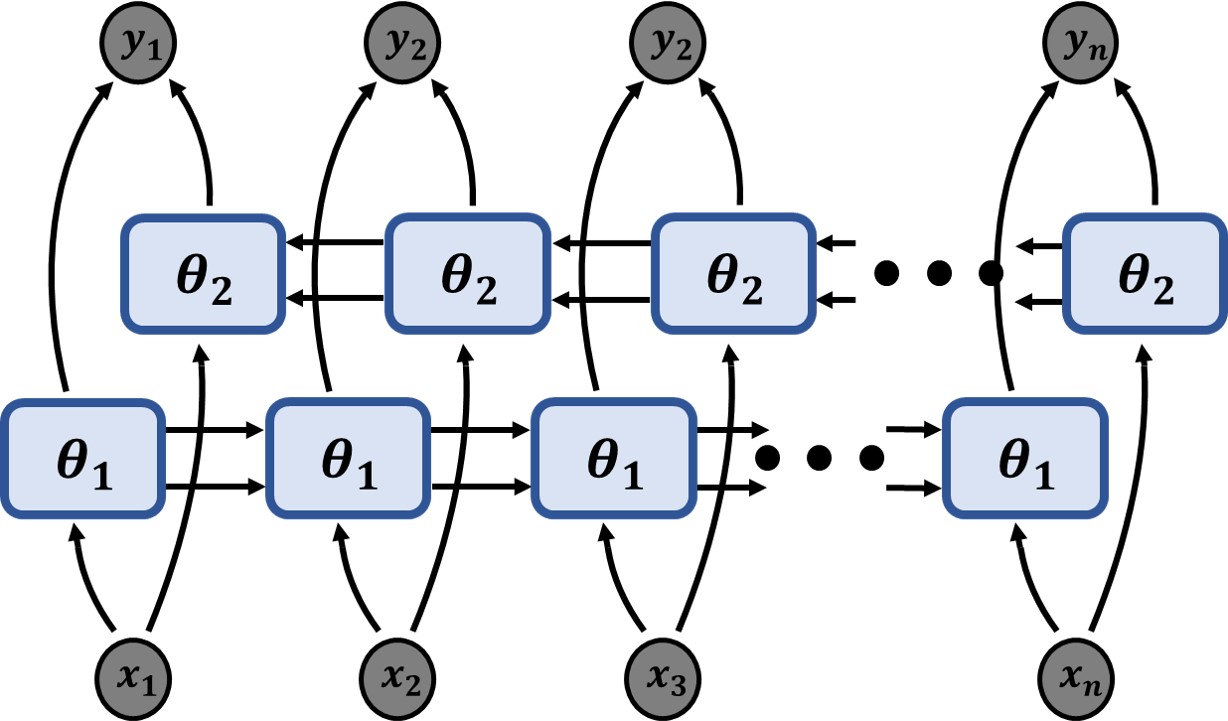}
    \caption{Single-layer bidirectional architecture, with LSTM units parametrised by $\theta_1$ and $\theta_2$.}
    \label{fig:BLSTM}
\end{minipage} 
\end{figure}

Recurrent Neural Networks (RNN) represent the quintessential DL model for sequence analysis.
RNNs recurrently apply a shared function over the inputs to learn temporal dependencies, but the basic RNN architecture is prone to vanishing or exploding gradients, which makes it difficult for the network to learn dependencies for long input sequences. Long short-term memory (LSTM) cells employ a gated recurrent unit with skip-connections, which allow gradients to flow for many time-steps, combating this problem \cite{hochreiter1997long}.
A general illustration of the LSTM block can be seen in Fig. \ref{fig:lstm_block}, comprised of input, forget and output gates. 
The forget gate allows the unit to restrain what information to be kept in the memory, $s$. 
The input gate encodes the current input, $x_t$, and controls its contribution to the memory, while the output allows the network to focus on relevant parts of the current state to compute the output, $h$. 

Since inputs are processed sequentially, the RNN architecture does not take future context into consideration when making current predictions. 
A bidirectional LSTM (BLSTM) could be used, where a second network simultaneously processes the input sequence in reverse, before outputs from both networks are used to compute the final outputs, as visualised in Fig. \ref{fig:BLSTM} \cite{schuster1997bidirectional}.

\subsection{Attention}\label{subsec_attention}
Learning to focus on relevant parts of the input, attention aims to allocate computational resources wisely and in some applications improve the interpretability of the models. 
Suppose an input sequence $x_1, x_2, ..., x_L \in \mathbb{R}^n$. 
Each input, $x_i$, is matched with all inputs by computing attention scores, using some shared attention mechanism, $a$, often a single- or multi-layered perceptron \cite{velivckovic2017graph, xu2015show}. 
A softmax function normalises the attention scores to generate probabilities:
\begin{equation}\label{eq:attn_2}
    \alpha_{ij} = \frac{\exp(a(x_i, x_j))}{\sum_{k=1}^L \exp(a(x_i, x_k))}.
\end{equation}
Finally, the updated values, $z_i \in \{z_1, z_2, ..., z_L\}$, are obtained by taking a weighted sum of the inputs, where a value function, $F$, is used to transform the inputs, along with a non-linearity, $\sigma$:
\begin{equation}\label{eq:attn_3}
    z_i = \sigma(\sum_{j=1}^L\alpha_{ij} F(x_j)).
\end{equation}
Multi-head attention has also been found beneficial for many applications, where the process in eq. (\ref{eq:attn_2}) and (\ref{eq:attn_3}) is repeated, and their outputs concatenated \cite{vaswani2017attention}. 

\subsection{Graph Neural Networks}\label{subsec_gnn}
Even though CNNs are excellent at analysing image data, a fundamental limitation of these models, is that these networks cannot naturally adapt to irregular domains.
Graph Neural Networks (GNN) aim to combat this by creating a more general framework to analyse problems thought to have important relational structures, but lack an inherent grid-like ordering.
The general concepts of GNNs will now be presented, based on the framework outlined in \cite{battaglia2018relational}.

\begin{figure}
\begin{minipage}[t]{0.39\linewidth}
        \centering
    \includegraphics[width=5cm]{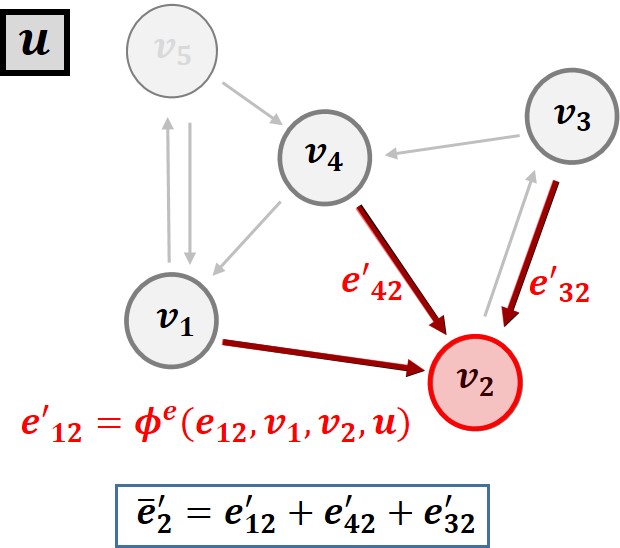}
    \caption{Visualisation of the $\rho ^{e\to v}$ operation for an arbitrary graph using summation aggregate function.}
    \label{fig:e2n_agg}
\end{minipage}%
    \hfill%
\begin{minipage}[t]{0.57\linewidth}
    \centering
    \includegraphics[width=7cm]{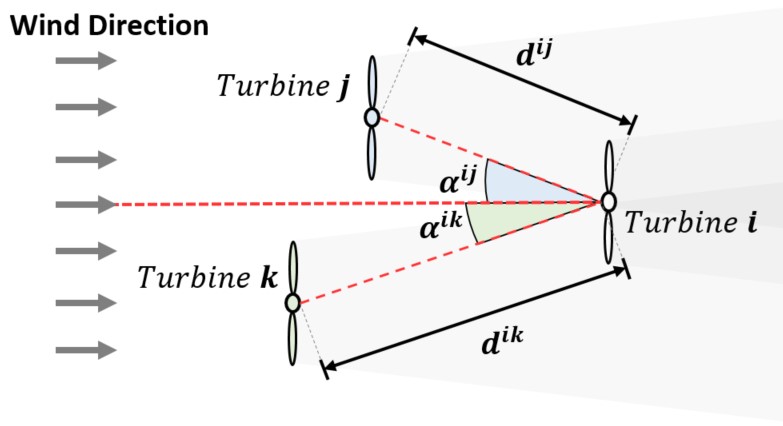}
    \caption{Visualisation of the two upstream neighbours for a turbine $i$ for a particular wind direction, with the relevant wake distances and angles shown.}
    \label{fig:upstrm_rep}
\end{minipage} 
\end{figure}

A graph can be defined as a tuple containing global, edge and node features, $G = (u, E, V)$.
$E$ and $V$ are the sets containing the edge and node features $e_{ij} \in E$ and $v_i \in V$. 
$v_i$ are attributes specific to a node $i$, while $e_{ij}$ are attributes associated with `sending' from node $i$ to $j$, such as the distance between two nodes.  
There is only a single global feature for each graph, stored in $u$.  
Similar to the stacking of layers in CNNs, GNNs employ a series of graph blocks which extract relevant features sequentially.
Global, edge and node features are updated using three functions within each block, to perform per-edge, per-node and global updates, as:
\begin{align}
    e'_{ij} &= \phi^e(e_{ij}, v_i, v_j, u) \\
    v'_j &= \phi^v(v_j, \bar{e}'_j, u) \\
    u' &= \phi^u(\bar{e}', \bar{v}', u) , 
\end{align}
where $\phi^{(\cdot)}$ represent the update functions, here assumed to be MLPs, $(\cdot)'$ indicate updated and $\bar{(\cdot)}$ aggregated features.
Updated features are fed as inputs to the next graph block.
Three aggregation functions map edges to nodes, edges to globals and nodes to globals as follows:
\begin{align}
    \bar{e}'_j &= \rho^{e\to v}(E'_j), \text{   where } E'_j = \{ e'_{ij} | \forall i \in R_j \} \\
    \bar{e}' &= \rho^{e \to u}(E'),    \text{   where } E' = \{ e'_{ij} | \forall (i,j) \in U \} \\
    \bar{v}' &= \rho^{v \to u}(V'),   \text{    where } V' = \{ v'_i | \forall i \in I \} .
\end{align}
$\rho^{(\cdot) \to (\cdot)}$ are the different aggregation operations, e.g. mean or sum. 
$R_j$, $U$ and $I$ are index sets containing indices for all nodes sending to node $j$, all edge indices $(i,j)$ and all node indices, respectively.
These sets define the connectivity of the graph.
A visualisation of an edge-to-node sum aggregation, $\rho ^{e\to v}$, can be seen in Fig. \ref{fig:e2n_agg}.

\section{Methodology}\label{sec_methodology}
\subsection{Data Description}\label{subsec_data}
The wind farm data was simulated in FLORIS \cite{FLORIS_2021}, using a Gaussian-based wake model \cite{king2020controls}. 
A range of random farm configurations were simulated for different wind speeds and directions, varying the layout and number of turbines from 28 to 150. 
18,280 different scenarios were simulated by varying the farm layout, number of turbines and wind conditions, resulting in  1,200,862 power values for individual turbines. 
All data was transformed by scaling features within $(0, 1)$, before being split into test, train and validation sets. 
The test data was obtained by randomly sampling 20\% of the original data. The remaining 80\% of the data was partitioned into training and validation sets, using the same 80-20\% split. 

\subsection{MLP and BLSTM Models}\label{subsec_annblstm}
The first two DL-based wake models were trained to predict the power production for individual turbines, with information on the locations of upstream neighbours and the wind speed.
Turbines that were within $\pm30^o$ (i.e. $|\alpha| < 30^o$ with reference to Fig. \ref{fig:upstrm_rep}) were considered upstream neighbours.
Depending on a turbine's location within a farm and the wind direction, its number of upstream neighbours would vary for different time-steps, resulting in variable length inputs.
It should be noted that even though we refer to `time' here, temporal correlations were not considered, but used to indicate the different inputs with distinct wind conditions and farm layouts. 
Considering for example a turbine, $i$, situated downstream of two turbines, $j$ and $k$, for a particular wind direction (depicted in Fig. \ref{fig:upstrm_rep}), the BLSTM model would look as: 
\begin{equation}\label{eq:blstm}
    \hat{P}^{(i)}_t = \textbf{h}(ws_t, \textbf{g}(d^{ij}, \sin(\alpha_t^{ij}), \cos(\alpha_t^{ij}), d^{ik}, \sin(\alpha_t^{ik}), \cos(\alpha_t^{ik}))),
\end{equation}
where $\hat{P}^{(i)}_t$ is the predicted power for turbine $i$ at time $t$, $ws_t$ the free stream wind velocity, $d^{ij}$ the distance between turbines $i$ and $j$, and $\alpha_t^{ij}$ the angle of $j$ relative to the wind direction. 
$\textbf{h}$ and $\textbf{g}$ are functions represented by MLP and BLSTM networks, respectively.
The MLP model is very similar to eq. (\ref{eq:blstm}), but with padded inputs to ensure constant length inputs and without the transform $\textbf{g}$ represented by the BLSTM network.

Additionally, two baseline models were used to compare against, where one would be trained to predict total farm power and another to predict power for individual turbines.
The inputs to these models only contained information on the total number of turbines in the farm and the free stream wind velocity. 
Since the baseline MLP models did not have any information on farm layout or upstream neighbours, these models should not be able to accurately predict wake losses. 
If the other models were able to achieve significantly lower prediction errors than the baselines, it would be concluded that they captured, at least part of, the wake losses.

\subsection{GNN Model with Attention}
\begin{figure}[b]
    \centering
    \includegraphics[width=10cm]{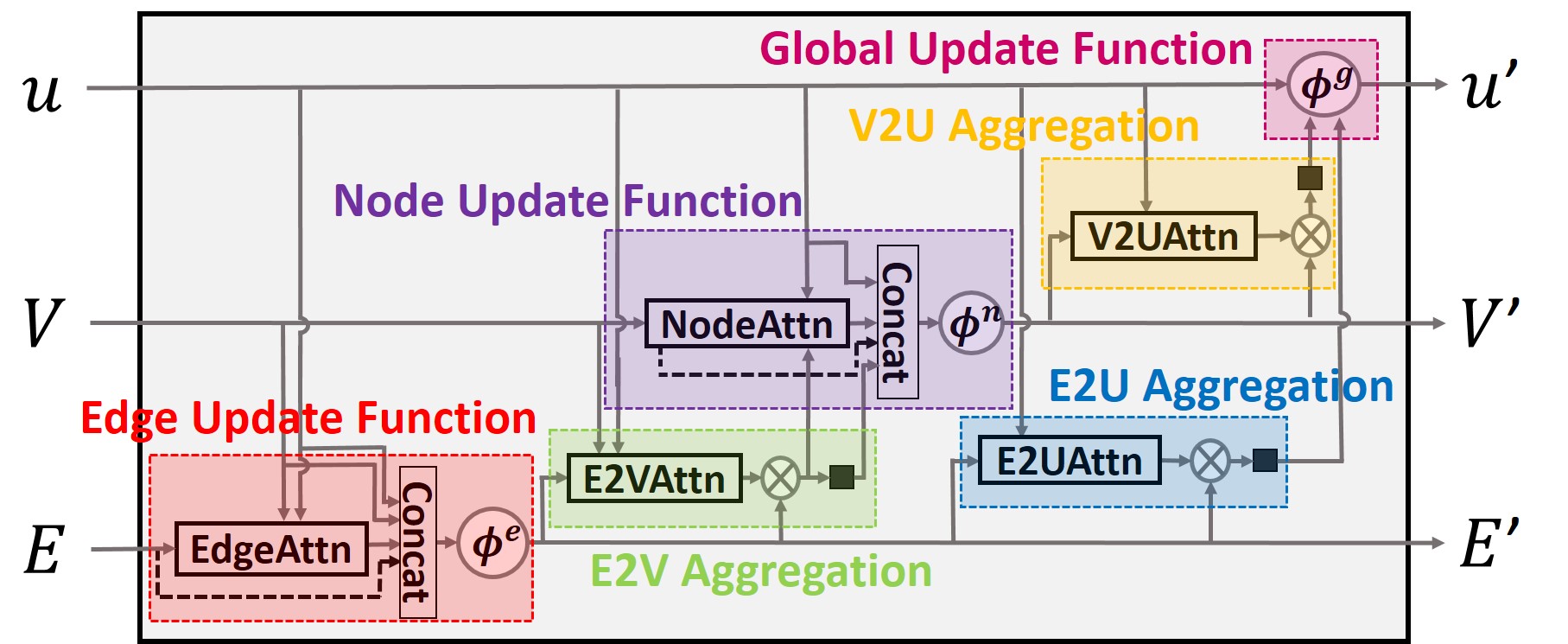}
    \captionof{figure}{The proposed attention-based graph block. To retrieve the original graph block, without attention, all attention blocks would be omitted, denoted by \textbf{-Attn}. \img{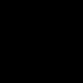}  and $\phi$ denote the aggregation operation and transforms represented by MLPs, respectively.}\label{fig:gnn_architecture}%
\end{figure}
For the GNN models, a graph was created for each wind condition and farm layout, where turbines are represented by nodes in the graph, with free-stream wind velocity, $ws_t$, as input node and global features. 
Interactions between nodes were encoded in the edges, with input features as; $d_{ij}$, $\sin(\alpha_t^{ij})$ and $\cos(\alpha_t^{ij})$.
An edge exists, sending from turbine $i$ to $j$, if turbine $i$ is upstream of turbine $j$.
The number of edges and connectivity of a graph would therefore change for different wind directions. 
The GNN produce updated graphs, trained to predict power for individual turbines embedded in the updated node features, and total farm power in the global. 

Graph Attention Networks (GAT) were first introduced by Veličković et al. \cite{velivckovic2017graph}, where attention was used to update nodes in a graph. 
Edge features might also be important for certain applications, and some studies have therefore tried to extend the concept of GATs to facilitate such features \cite{chen2021edge}. 
Incorporating attention into the model, our first aim was to be able to elegantly handle a variable number of upstream neighbours.
Secondly, the attention weights can be seen as the relative importance placed on different parts of the input by the network. 
For this particular application, the hope was that the network would learn weights to focus on the most influential upstream turbines to determine wake losses, which should hopefully align with some physical intuition about wind farm dynamics and enhance interpretability. 

The proposed method is made up of a modular architecture, where attention can be applied to the different update or aggregate operations outlined in Sec. \ref{subsec_gnn}.
A visualisation of the proposed graph block can be seen in Fig. \ref{fig:gnn_architecture}, where the original architecture without attention can be retrieved if MLPs and desired aggregate functions are used in place of the highlighted update and aggregation boxes.
Depending on the application, the framework allows for various graph attention architectures to be tested and attention to be applied to the relevant parts of the network for a particular task.
For instance, if attention was only to be applied in the node update function in Fig. \ref{fig:gnn_architecture}, one would retrieve a similar architecture as the GAT \cite{velivckovic2017graph}.

Considering the \textbf{NodeAttn} block in Fig. \ref{fig:gnn_architecture}, the network will attend to all nodes sending to a target node. 
For a turbine $j$, this means computing weights for all $i \in R_j$. 
Such an operation can be considered as masked attention, since the model does not compute attention weights over the entire input, but only for the nodes that are in the neighbourhood of $j$. 
For this particular study, it was decided to use scaled dot product attention, as in \cite{vaswani2017attention}:
\begin{equation}\label{eq:gat_attn_sa}
    \alpha_{ij} = \frac{\exp((K^T[u \mathbin\Vert v_{i} \mathbin\Vert e'_{ij}])^T Q^T[\bar{e}'_j\mathbin\Vert v_{j}]/\sqrt{d})}{\sum_{\forall k \in R_j} \exp((K^T[u \mathbin\Vert v_{k} \mathbin\Vert e'_{kj}])^T Q^T[\bar{e}'_j \mathbin\Vert v_{j}]/\sqrt{d})},
\end{equation}
where $\mathbin\Vert$ denotes the concatenation operator. $Q$ and $K$ are weights for linear transformations to compute $d$-dimensional keys and queries.
However, it should be noted that instead of using the keys and queries mapping in scaled dot product attention, an MLP or other function could still be used, as in eq.(\ref{eq:attn_2}).
The outputs from the \textbf{NodeAttn} block, $z_j$, were then computed as
\begin{equation}
    z_j = \sigma(\sum_{\forall i \in R_j} \alpha_{ij}F^Tv_i).
\end{equation}

\begin{figure}
    \parbox[t]{9cm}{\null
      \centering
      \includegraphics[width=8cm]{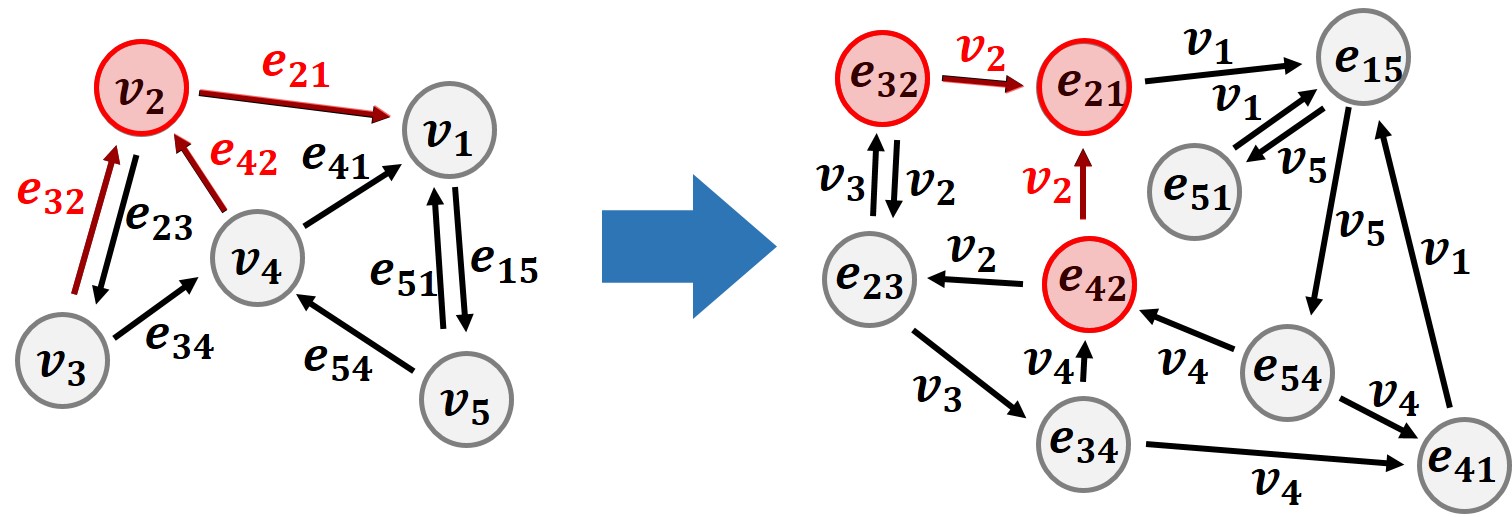}%
      \captionof{figure}{Edge-transformed graph representation. The same operations as in the \textbf{NodeAttn}, could then be applied in the \textbf{EdgeAttn} in Fig.\ref{fig:gnn_architecture}.}\label{fig:edge_graph}%
    }
    \parbox[t]{6cm}{\null
    \centering
      \vskip-\abovecaptionskip
      \captionof{table}[t]{MAE Test Results.}\label{tab::results}%
      \vskip\abovecaptionskip
      \resizebox{7cm}{!}{
      \begin{tabular}{lcc} 
        \hline
        Model & MAE Turbine & MAE Farm \\
        \hline
        BS\_Farm & - & 0.015820 \\
        BS\_Turb & 0.052266 & - \\
        MLP & 0.005652 & 0.000790 \\
        BLSTM & 0.005331  & 0.000720 \\
        O-Graph & \textbf{0.001426} & 0.000511 \\
        N-Graph & 0.001452 & 0.000526 \\
        F-Graph & 0.001441 & \textbf{0.000495} \\
        \hline
        \end{tabular}}
    }
\end{figure}

For the \textbf{EdgeAttn}, the same operations were performed, but now attending over the edges that send to a target edge.
To obtain a representation where edges send to other edges via nodes, the input graph was transformed to a temporary edge-graph, where an edge $e_{ij}$ sends to $e_{jk}$ via the node $v_j$.
The edge-graph mapping was inspired by the works in \cite{chen2021edge}, producing a new graph with edge features mapped to nodes.
Different to \cite{chen2021edge}, the proposed mapping works for directed graphs, as visualised for an arbitrary graph in Fig. \ref{fig:edge_graph}. 
The same attention mechanism used for the \textbf{NodeAttn} could then be used for the \textbf{EdgeAttn} operation. 

Considering the attention operations for the aggregations, similar operations were used, but with different neighbourhoods. 
For the \textbf{E2VAttn}, masked attention was performed over all edges sending to the same node.
Since global features are shared across a single graph, \textbf{V2UAttn} and \textbf{E2UAttn} perform attention over the full input, i.e. not masked, using all nodes or edges. 

\section{Results and Discussion}\label{sec:results}
\subsection{Wind Power Prediction Accuracy}
To assess the performance of the proposed models in predicting wind power production, the Mean Absolute Error (MAE) metric was used:
\begin{equation}
    \text{MAE} = \frac{1}{T}\sum_{t=1}^T |\hat{y}_t - y_t|,
\end{equation}
where $y$ are true and $\hat{y}$ predicted values, for $T$ samples. 
Two losses were computed for predicting individual turbine and total farm powers, denoted "MAE Turbine" and "MAE Farm" in Table \ref{tab::results}. 
The MLP and BLSTM models improved the accuracy ten- and twenty-fold for individual turbine and farm predictions, respectively, compared to the basline models, denoted "BS\_" in Table \ref{tab::results}.
The BLSTM outperformed the MLP model, with an approximate $5$ and $10$\% improvement for the turbine and farm predictions. 
Besides a slightly inferior performance, a drawback of the MLP model was that it would not be able to make predictions on wind farms with a greater number of turbines than what was present in the training data. 
Nevertheless, comparing the models to the baselines, it was concluded that both architectures were able to capture wake losses, though the BLSTM seemed potentially better suited for the variable length inputs.

Considering the graph models in Table \ref{tab::results}, all achieved significantly better prediction performance than the other models. 
Here, O-Graph is the vanilla GNN model without attention. 
F-Graph employ all attention modules in Fig. \ref{fig:gnn_architecture} for the first graph block layer, while the N-Graph only use attention for the edge-to-node aggregation and the node-update function.
The reason for including both the F- and N-Graph models, was to demonstrate the modularity of the proposed graph attention framework.
The O-Graph model reported the smallest MAE of 0.00143 for individual turbine predictions, while the F-Graph yielded the best results for total farm predictions, with an MAE of 0.000495.
The prediction performance of the N-Graph model was close to the other graph models, with MAEs of 0.00145 and 0.000526. 
Despite slightly different prediction performances between the three graph models, we argue that the comparable results demonstrate the functioning of the proposed  framework in Fig. \ref{fig:gnn_architecture}.

\subsection{Analysis of Attention Weights}\label{subsec::attn_weights}
\begin{figure}[t]
    \centering
    \includegraphics[width=15cm]{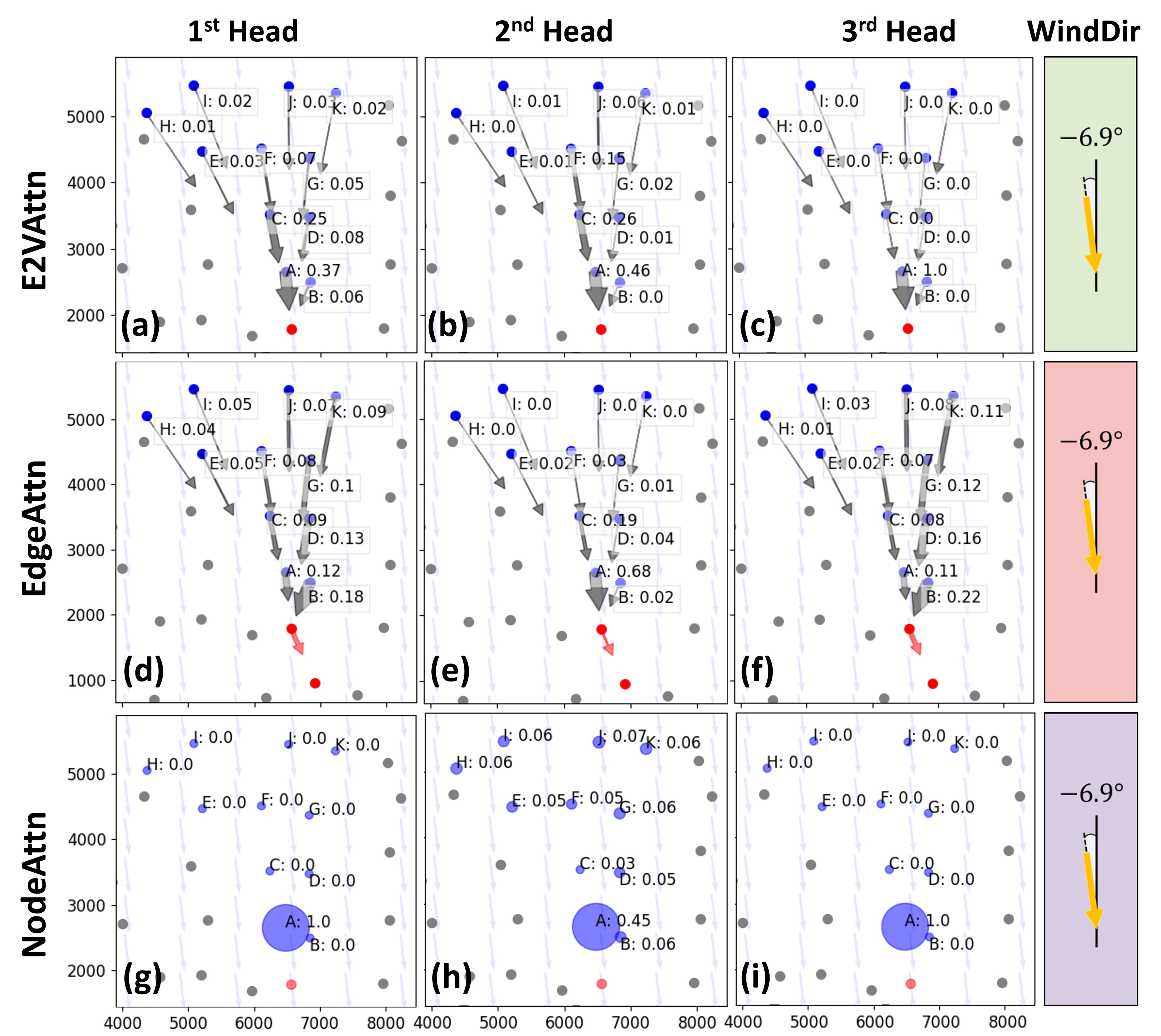}
    \caption{Attention weights from the F-Graph model for a randomly chosen farm configuration and wind condition. Red colouring represents a target wind turbine or edge, while blue represents upstream neighbours. \textbf{E2VAttn}, \textbf{EdgeAttn} and \textbf{NodeAttn} refers to the blocks in Fig. \ref{fig:gnn_architecture}.}
    \label{fig:attn_weights_1}
\end{figure}

\begin{figure}[t]   
    \centering
    \includegraphics[width=15cm]{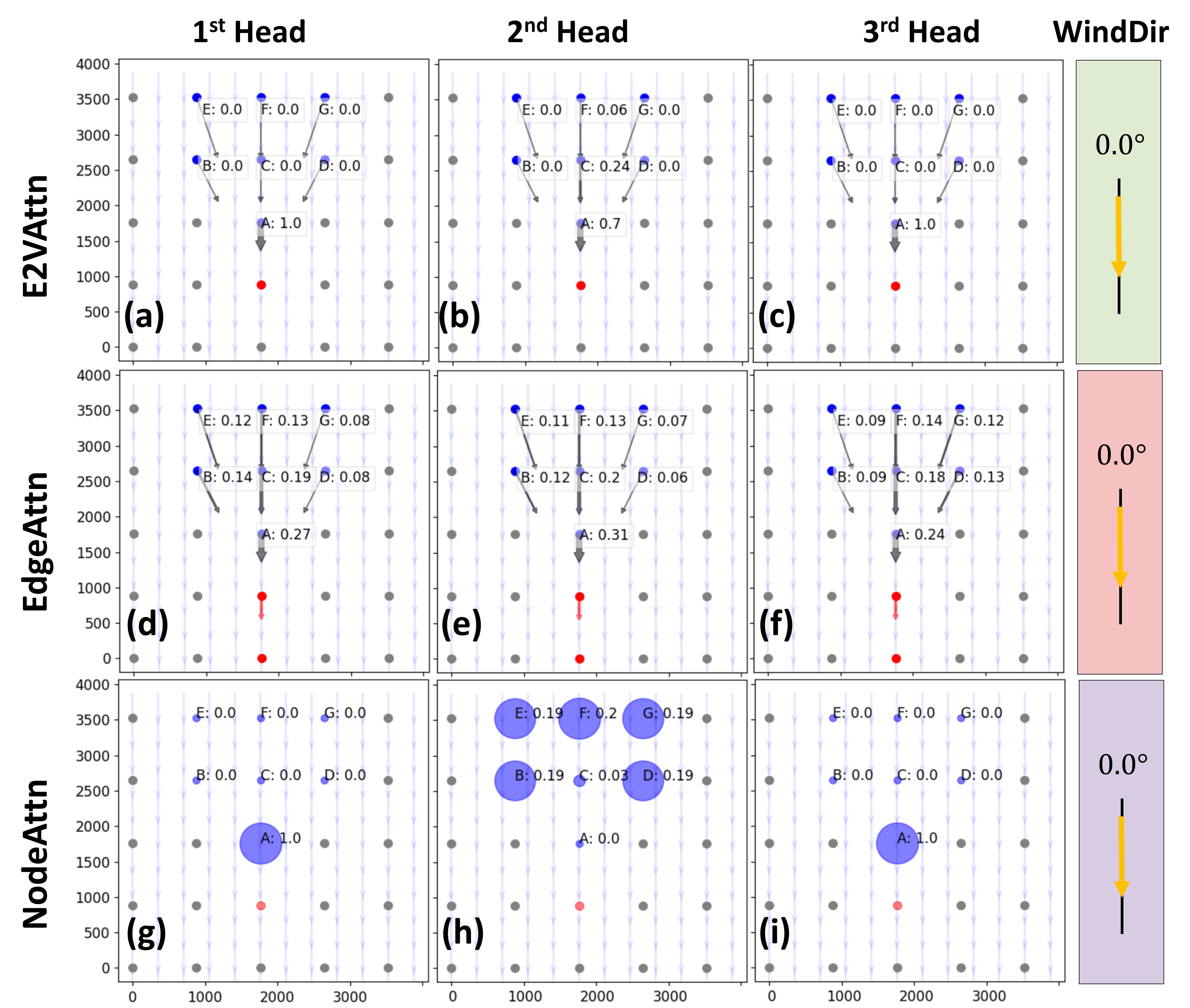}
    \caption{Attention weights from the F-Graph model for an ordered wind farm layout with 100\% wake overlap. Red colouring represents a target wind turbine or edge, while blue represents upstream neighbours. \textbf{E2VAttn}, \textbf{EdgeAttn} and \textbf{NodeAttn} refers to the blocks in Fig. \ref{fig:gnn_architecture}.}
    \label{fig:attn_weights_2}
\end{figure}

Each of the five attention operations in the F-Graph model used three heads, computing three sets of attention weights for each operation. 
Fig. \ref{fig:attn_weights_1} shows the attention weights from the edge-to-node aggregation, edge- and node-update functions, for a randomly selected farm layout and wind condition during testing, while global attention weights were excluded from the plots due to the size of the farm.
Furthermore, another scenario was also investigated, shown in Fig. \ref{fig:attn_weights_2}, in which the turbines were situated in a two-dimensional grid, with the wind direction set to zero degrees. 
The dots in the figures represent turbine locations, with red being the target and blue its upstream neighbours for the particular wind conditions. 
Arrows represent edges, sending from the upstream turbines to the target. 
The width of the arrows and size of the circles visualise the relative weighting computed by the \textbf{E2VAttn}, \textbf{EdgeAttn} and \textbf{NodeAttn}, respectively, given by the corresponding numbers in Fig. \ref{fig:attn_weights_1} and \ref{fig:attn_weights_2}. 

Considering the edge-to-node attention weights, it was seen that the heads seemed to focus on slightly different aspects of the input. The main focus was placed on the edge sending from the turbine directly upstream and in close vicinity to the target, i.e. the edge sending from turbine A in Fig. \ref{fig:attn_weights_1} and \ref{fig:attn_weights_2} a-c.
From Fig. \ref{fig:attn_weights_1}, it was evident that the most important edges were not necessarily those sending from the turbines closest to the target, as hardly any weighting was placed on the edge from turbine B.
This was interesting, as it showed that the networks were able to learn more complex behaviours, than simply how close turbines were to the target. 
Looking at the second head \textbf{E2VAttn}, incrementally lower attention was placed on edges moving directly upstream from the target, as 0.46, 0.26 and 0.15, for edges A, C and F in Fig. \ref{fig:attn_weights_1}. Hardly any weighting was placed on other edges, confirmed by the weights in Fig. \ref{fig:attn_weights_2}.
Intuitively, the second head might be thought to help determine wake losses by capturing changes when moving directly upstream from the target wind turbine.

Moving to the \textbf{EdgeAttn}, the second head weights showed similar patterns, with weights tapering off when moving upstream from the sending target turbine. 
Since this attention mechanism was used to update edges, there was effectively two target turbines, one sending and one receiving. 
Looking at the first and third head weights in plots d and f of Fig. \ref{fig:attn_weights_1}, weights were more evenly distributed, but still with the most focus placed on turbines upstream of the target edge. 
Different to the \textbf{E2VAttn}, more attention was placed on edges sending from turbines moving clock-wise away from the upstream line, i.e. B, D, G and K.
In particular, while the edge from turbine B was negligible for the \textbf{E2VAttn}, it was highly significant for the \textbf{EdgeAttn} in Fig. \ref{fig:attn_weights_1}.
This behaviour was likely due to the location of the receiving target turbine, with the turbines receiving significantly more weighting since they were influential to the power of the receiving target turbine. 
Observing that the location of the receiving turbine was important and the learned behaviour for the first two attention networks were different was interesting, as it showed that the network could learn complicated structures in the data.

Again, attention was mainly placed on the turbine directly upstream of the target for the \textbf{NodeAttn} weights. 
Two of the heads placed the entire weighting on this turbine, while the second head also attended to the other turbines. 
For the ordered farm layout in Fig. \ref{fig:attn_weights_2}, the second head did not place any weighting on turbine A.
No difference was evident from the first and third head weights, which could imply two things;
either, that one head focuses on dependencies that were not present for the particular layouts and wind conditions in Fig. \ref{fig:attn_weights_1} and \ref{fig:attn_weights_2}, or that one of the three heads for the \textbf{NodeAttn} were redundant. 
The distributed weighting for the second head might indicate that an averaging of the features for upstream turbines provides important information to determine the power production for a target turbine.

Even though the attention-based models were not able to significantly improve on MAEs, investigating the attention weights showed that the proposed modules were able to learn meaningful dependencies in the data. 
An important finding, was that the attention mechanisms did not learn a simple average of all the input features, as this would mean that the mechanisms would replicate a similar behaviour to the O-Graph model.
Although we were able to deduce some physical intuition about what the F-Graph model learned, the conclusions drawn were quite simple. 
Nevertheless, one might expect that attention mechanisms could be used more extensively to aid in applications where we want to learn about and understand more complicated physical properties or patterns, that are normally inaccessible to the users of DL models. 

Generally, it was observed that the model seemed exceedingly confident, often placing almost all the weighing on a small subset of the inputs, entirely disregarding others. 
For the \textbf{E2VAttn} weights, this was evident, as turbine B received almost no weighting in Fig. \ref{fig:attn_weights_1}. 
One could therefore argue that the model might be too sensitive, as slight offsets in the inputs could result in the model focusing on the wrong features.
Even though the high confidence allowed for perspicuous interpretations of the results, it was thought that this was an inherent reflection of the data used. 
Wind farm data simulated in FLORIS lack uncertainty, which would be a prominent feature of any real dataset, due to sensor drifts, non-stationary atmosphere conditions and non-uniform wind field across a farm. 
If the models were instead trained on real site measurements, it is likely that attention weights might be more evenly distributed, with less values close to 0 or 1, as information from only a small number of upstream turbines might be a poor reflection of global characteristics.
For future research, it would be particularly interesting to investigate how the models operate when trained on real farm measurements or data simulated using more accurate numerical methods that considers non-stationary atmosphere.
Utilising real datasets, it is also thought that the superior performance of the GNN models, with regards to MAEs, might slightly curtail, as other features than interaction losses might become more pronounced. 
Nevertheless, for this preliminary study, the rapid simulations and clean data obtained using FLORIS was important as it allowed for better analysis of the attention weights, enabling investigation into whether the proposed framework behaved as desired.


\section{Conclusion}\label{sec:conclusion}
In this study, it was seen that graph models can outperform other DL architectures in predicting turbine powers using data simulated in FLORIS.
A novel framework was proposed to apply attention to all the relevant update- or aggregate-functions in a GNN model and the framework achieved results on-par with vanilla GNN models. 
Through analysis of the attention weights, it was seen how the networks learned interactions between turbines and were able to produce results that aligned well with some physical intuition about wake losses.
For future research, it will be interesting to investigate how attention mechanisms can help to learn more complex physical behaviours or improve prediction performance of graph models applied to wind-based technologies.
Furthermore, the modular framework allows full flexibility for the user and it is a hope that it can aid in the development of graph-attention models for a range of different research applications in the future.

\section*{Acknowledgements}
This work was in part financed by the research project ELOGOW (Electrification of Oil and Gas Installation by Offshore Wind). ELOGOW (project nr: 308838) is funded by the Research Council of Norway under the PETROMAKS2 program with financial support of Equinor Energy AS, ConocoPhillips Skandinavia AS and Aibel AS.
The authors would also like to thank the developers of FLORIS \cite{FLORIS_2021} for enabling wind farm simulations used for this study.

\bibliographystyle{unsrt}
\bibliography{ref}

\end{document}